\documentclass[runningheads]{llncs}

\usepackage{amsfonts,amssymb,amstext}
\usepackage{mathtools}
\usepackage{booktabs}
\usepackage{longtable}
\usepackage{nicefrac}
\usepackage{xspace}
\usepackage{url}
\usepackage{graphics,color}
\usepackage[algo2e,ruled,vlined,linesnumbered]{algorithm2e}
\SetKwRepeat{Do}{do}{while}%
\usepackage{algorithmicx}
\usepackage{algorithm}
\usepackage[noend]{algpseudocode}

\usepackage{graphicx}
\usepackage{wrapfig}
\usepackage{cite}
\usepackage{hyperref}

\usepackage{pgfplots}
\usepackage{pgfplotstable}
\usetikzlibrary{spy,external}
\usepackage{subfig}
\usepackage{longtable}

\newcommand{\ignore}[1]{}

\newcommand{\highlight}[1]{\textcolor{blue}{#1}}

\allowdisplaybreaks

\newcommand{\R}{\mathbb{R}}

\renewcommand{\epsilon}{\varepsilon}

\DeclareMathOperator*{\argmax}{arg\,max}

\newcommand{\onemax}{\textsc{OneMax}\xspace}

\newcommand{\OneMax}{\textsc{OneMax}\xspace}
\newcommand{\OM}{\textsc{Om}\xspace}
\newcommand{\LeadingOnes}{\textsc{LeadingOnes}\xspace}
\newcommand{\leadingones}{\textsc{LeadingOnes}\xspace}
\newcommand{\LO}{\textsc{Lo}\xspace}

\newcommand{\Plateau}{\textsc{Plateau}\xspace}
\newcommand{\Ruggedness}{\textsc{Ruggedness}\xspace}
\newcommand{\Neutrality}{\textsc{Neutrality}\xspace}

\newcommand{\oplab}{$(1 + \lambda) \text{ EA}(A,b)$\xspace}

\newcommand{\opl}{$(1 + \lambda)$~EA\xspace}
\newcommand{\lea}{$(1 + \lambda)$~EA\xspace}
\newcommand{\qea}{$(1 + \lambda)$~QEA\xspace}
\newcommand{\hqea}{$(1 + \lambda)$~HQEA\xspace}

\newcommand{\oplQ}{\qea}

\newcommand{\hybrid}{\hqea}

\newcommand{\AB}{\oplab}

\newcommand{\tworate}{$(1+\lambda) \text{ EA}_{r/2,2r}$\xspace}

\newcommand{\rate}{\tworate}

\pgfplotscreateplotcyclelist{colorcycle}{%
red!80!black,mark=*,mark size=1\\
blue,mark=square*,mark size=1\\
cyan!80!black,mark=*,mark size=1\\
violet!80!black,mark=square*,mark size=1\\
orange,mark=diamond*\\
}

\pgfplotscreateplotcyclelist{noqcolorcycle}{%
red!80!black,mark=*,mark size=1\\
blue,mark=square*,mark size=1\\
cyan!80!black,mark=*,mark size=1\\
orange,mark=diamond*\\
}

\DeclareMathOperator{\mutate}{mutate}

\usepackage{fullpage}

\begin{document} 
\title{Hybridizing the 1/5-th Success Rule with Q-Learning for Controlling the Mutation Rate of an Evolutionary Algorithm} 
\titlerunning{Hybridizing the 1/5-th Success Rule with Q-Learning} 

\author{Arina Buzdalova\inst{1} \and
Carola Doerr\inst{2} \and
Anna Rodionova\inst{1}}

\authorrunning{A. Buzdalova, C. Doerr, A. Rodionova}

\institute{
	ITMO University, 49 Kronverkskiy ave., Saint Petersburg, Russia, 197101\\ 	
	\email{abuzdalova@gmail.com}	
	\and
	Sorbonne Universit{\'e}, CNRS, LIP6, Paris, France\\
	\email{Carola.Doerr@lip6.fr}
}

\maketitle

\begin{abstract} 
It is well known that evolutionary algorithms (EAs) achieve peak performance only when their parameters are suitably tuned to the given problem. Even more, it is known that the best parameter values can change during the optimization process. Parameter control mechanisms are techniques developed to identify and to track these values. 

Recently, a series of rigorous theoretical works confirmed the superiority of several parameter control techniques over EAs with best possible static parameters. Among these results are examples for controlling the mutation rate of the $(1+\lambda)$~EA when optimizing the OneMax problem. However, it was shown in [Rodionova et~al., GECCO'19] that the quality of these techniques strongly depends on the offspring population size $\lambda$.

We introduce in this work a new hybrid parameter control technique, which combines the well-known one-fifth success rule with Q-learning. We demonstrate that our HQL mechanism achieves equal or superior performance to all techniques tested in [Rodionova et~al., GECCO'19] and this -- in contrast to previous parameter control methods -- simultaneously for all offspring population sizes $\lambda$. We also show that the promising performance of HQL is not restricted to OneMax, but extends to several other benchmark problems.  

\end{abstract}

\sloppy{ 
\section{Introduction}

The problem of selecting suitable parameter configurations for an evolutionary algorithm is frequently considered to be one of the most essential drawbacks of evolutionary computation methods, and possibly a major obstacle towards wider application of these optimization techniques in practice~\cite{LoboLM07}. 

Automated configuration techniques such as~SPOT~\cite{SPOT}, irace~\cite{irace}, SMAC~\cite{SMAC}, hyperband~\cite{li2016hyperband}, MIP-EGO~\cite{MIP-EGO}, BOHB~\cite{BOHB}, and many others have been developed to assist the user in the decisive task of selecting suitable parameter configurations. These \emph{parameter tuning} methods, however, require to test different parameter combinations before presenting a recommendation. They are therefore rather time-consuming, and are not applicable when the possibility for such training is not given, e.g., when the problem is truly black-box, with no/only little information about its fitness landscape structure. 

An orthogonal approach to solve the algorithm configuration problem is \emph{parameter control}, which does not require a priori training, and aims at identifying suitable parameter combinations \emph{on the fly}, i.e., while executing the optimization~\cite{EibenHM99,KarafotiasHE15,LoboLM07}.  
Apart from being more generally applicable than parameter tuning, parameter control also bears the advantage of being able to adjust the search behavior of the evolutionary algorithm to the different stages of the optimization process. Most state of the art evolutionary algorithms therefore make use of parameter control, in particular in the continuous domain, where a decreasing search radius is needed to eventually converge towards an optimal point. However, one should not forget that parameter control mechanisms, too, introduce their own hyperparameters, which need to be adequately set by the user prior to running the algorithm. Here again one can apply parameter tuning (e.g., via so-called per-instance algorithm configuration~\cite{BelkhirDSS17}),  
but the general hope is that the setting of the hyperparameters is less critical to achieve reasonable performance. 

However, while parameter control is routinely used in numerical optimization, its potential remains far from being well exploited in the optimization of problems with discrete decision variables, where it has only recently re-gained momentum as a now very active area of research. In particular in the sub-domain of runtime analysis, parameter control has enjoyed rising attention in the last years, as summarized in~\cite{DoerrD20survey}. 

A particularly well-researched topic in the theory literature for parameter control in discrete optimization heuristics is the $(1+\lambda)$~Evolutionary Algorithm (EA) with dynamic mutation rates and fixed offspring population size $\lambda$ optimizing the \onemax problem (the problem of controlling $\lambda$ has also been addressed, e.g., in~\cite{LassigS11}, but has received much less attention so far). Not only was this problem one of the first ones for which dynamic mutation schemes were approximated~\cite{Back92}, and not only is it frequently used as a test case for empirical works~\cite{DaCostaGECCO08}, but 
it is also one of the few problems for which we have a very solid theoretical understanding. 

Extending the previous work from~\cite{DoerrYRWB18}, we have presented at GECCO'19 a comparative empirical study of several mechanisms suggested in the theory literature~\cite{Rodionova-GECCO}. Among other findings, we demonstrated that the efficiency of all benchmarked techniques depends to a large extent on the offspring population size $\lambda$. For example, we observed that  
the 2-rate \opl suggested in~\cite{DoerrGWY19} is the best among the tested algorithms when $\lambda$ is smaller than 50.  
For larger offspring population sizes, however, this algorithm is outperformed by a \opl which uses the one-fifth success rule to control the mutation rate. We also observed in~\cite{Rodionova-GECCO} that the ranking of the algorithms was identical for all tested dimensions $n \in [10^4..10^5]$.

\paragraph{Our Results.} 
The results presented in~\cite{Rodionova-GECCO} raise the question if one can achieve stable performance across all offspring population sizes $\lambda$. We address this problem by introducing a new parameter control scheme, which hybridizes the one-fifth success rule with Q-learning. 
More precisely, we first introduce the \qea, which uses Q-learning only to control the mutation rate. The \qea learns for each optimization state whether it should increase or decrease the current mutation rate (we use constant factor changes). We show that the \qea performs efficiently on \OneMax for all observed values of $\lambda$ when an appropriate lower bound $p_{\min}$ for the mutation rate is used. In absence of a well-tuned lower bound, however, the performance of the \qea drops significantly. We show that this dependence on the value of $p_{\min}$ can be mitigated by a hybridization of the \qea with the one-fifth success rule. More precisely, the hybrid Q-learning EA (the \hqea) extends the \qea by using the one-fifth success rule in states that have not been visited before and for those for which the \qea is ambiguous with respect to the two available actions. 

We show that, on \OneMax, the \hqea outperforms or at least performs on par with all algorithms tested in~\cite{Rodionova-GECCO}, and this simultaneously for all tested values of $\lambda \in [1..2^{12}]$ and also for both considered lower bounds for the mutation rate, $p_{\min}=1/n$ and $p_{\min}=1/n^2$, respectively. It therefore solves the issue of the other control mechanisms previously suggested in the theory literature. Note here that we do not have a theoretical convergence analysis of the \hqea. Given its complexity, it may be beyond the current state of the art in runtime analysis, as it requires to keep track of multiple states, which are highly dependent. We are nevertheless confident that the robust performance of the \hqea encourages further work on learning-based parameter control, and their hybridization with other classical control methods.    

In the last parts of this paper we also show that the promising performance of the \hqea is not restricted to \onemax. More precisely, we show that it performs well also on the \leadingones function, as well as on several benchmark functions suggested in~\cite{DoerrYHWSB20}. 

\paragraph{Related Work.} 
We are not the first to use reinforcement learning (RL) as a parameter control technique. An exhaustive survey of RL-based parameter control approaches can be found in~\cite{KarafotiasHE15}. Particularly, there are parameter control approaches based on techniques for the \emph{Multi-Armed Bandit Problem (MAB)}, see~\cite{FialhoCSS10} (and references mentioned therein) and~\cite{DoerrDY16PPSN} for a theoretical investigation of MAB-based parameter control.

In many of the known approaches, RL algorithms are used to select the parameter values directly. For numerical parameters, however, most common techniques require to either discretize the value space~\cite{karafotias-gecco} or to make use of quite sophisticated techniques~\cite{earpc,rost-petrova-parameter-selection-gecco16,ea-rl}, which are rather difficult to grasp without expert knowledge.

In contrast to such a direct selection of the parameter values, we use in this work an indirect approach which uses as actions the possibility to increase the current parameter value by some fixed multiplier, or decrease it. As we shall see below, this yields a simple, yet efficient, control mechanism. Like most common parameter control techniques, including those studied in this work, this indirect approach has the advantage of a smoother transition of the mutation rates between consecutive iterations. This behavior is beneficial if the optimal parameter values do not change abruptly, which is the case in many problems analyzed in theoretical works~\cite{Doerr19domi,DoerrDY16}, but also the case in many applications of evolutionary algorithms to machine learning problems, including hyperparameter optimization itself~\cite{Hoos18PPSN}. Exceptions to this rule exist, of course, and the jump functions~\cite{DrosteJW02} are a classical example for a problem requiring such an abrupt change. In such cases it may take the the parameter control mechanisms some time to adjust the mutation rate to the appropriate scale.

We note that a similar indirect control approach has been described in~\cite{rl-for-ea-adjusting}, where an indirect control of the step size of the (1+1) evolution strategy (ES) is described. In contrast to our work, however, this approach (which uses SARSA -- another common reinforcement learning algorithm -- instead of Q-learning) did not manage to outperform the (1+1)~ES with suitably tuned static step sizes. 
\section{Previous $(1+\lambda)$~EAs with Dynamic Mutation Rates}
\label{sec:opl}

We briefly review the algorithms studied in~\cite{Rodionova-GECCO} and summarize their main findings. We assume in our presentation that the algorithms operate on a problem $f:\{0,1\}^n \to \R$, with the objective to maximize this function. 

\textbf{The \opl.} The standard \opl is an elitist algorithm, which always keeps a current best solution $x$ in its memory. The \opl is initialized with a point chosen from the search space $\{0,1\}^n$ uniformly at random. In each iteration, $\lambda$ \emph{offspring} are sampled by applying standard bit mutation to the \emph{parent} $x$, i.e., the algorithm creates $\lambda$ offspring $y^{(1)}, \ldots, y^{(\lambda)}$ by creating $\lambda$ copies of $x$ and flipping each bit in these copies with some probability $0< p < 1$. The variable $p$ is commonly referred to as the \emph{mutation rate}. We set it to $p=1/n$ in our experiments, which is a standard recommendation and often a fall-back value if no indication is given that larger values could be beneficial. 
The best of the $\lambda$ offspring (ties broken uniformly at random) replaces the parent if it is at least as good. 
The \opl continues until some user-defined termination criterion is met (see ``implementation details'' below for our setting).   

\textbf{The \oplab.} The \oplab extends the \opl by an adaptive choice of the mutation rate $p$. Its (1+1) variant was suggested in~\cite{DoerrW18}, and we use a straightforward extension to the \opl by updating the mutation rate $p$ by $Ap$ if the best of the $\lambda$ offspring is at least as good as the parent and by decreasing the mutation rate to $bp$ otherwise. 
It is ensured that the mutation rate does not fall below some minimal mutation rate $p_{\min}>0$ and that it does not exceed $p_{\max}=1/2$, by capping the value of $p$ appropriately where required. As argued in~\cite{DoerrDL19}, this update rule is essentially a one-fifth success rule, even if this term was not mentioned in~\cite{DoerrW18}. The one-fifth success rule was originally suggested in~\cite{es-rechenberg,Devroye72,SchumerS68} and its interpretation for the discrete optimization is due to~\cite{KernMHBOK04}. More precisely, the idea is that the mutation rate should remain constant if a certain ratio of iterations is successful (i.e., produces a solution of better than previous-best quality). In our work, this success ratio is 1/2, whereas the traditional rule suggests a success ratio of $1/5$.

The \oplab has three hyperparameters, $A$, $b$, and $p_{\min}$. In our experiments, we set $A=2$, $b=1/2$, and consider $p_{\min} \in \{1/n, 1/n^2\}$. We initialize $p$ by $1/n$. Note that these values are not specifically tuned, but we chose them to be consistent with previous works, and in particular with~\cite{Rodionova-GECCO}. The reader interested in the sensitivity of the performance of the \oplab with respect to these parameters is referred to~\cite{DoerrW18} and~\cite{DoerrDL19} for an empirical and a theoretical investigation, respectively.

\textbf{The 2-rate \tworate.} The \tworate suggested in~\cite{DoerrGWY19} uses two different mutation rates in each iteration: half the offspring are created with mutation rate $p/2$ and the other $\lambda/2$ offspring are sampled with mutation rate $2p$. The mutation rate is parametrized as $p=r/n$ in the \tworate. The value of $r$ is updated after each iteration by a random decision which gives preference to the rate by which the best offspring has been created. The latter is selected with probability $3/4$, whereas the other one of the two tested mutation rates is chosen with probability $1/4$. As in the \oplab, the mutation rate is capped at $p_{\min} \in \{1/n^2, 1/n\}$ and $p_{\max}=1/2$, respectively.   

\paragraph{Implementation details.} We briefly summarize a few common assumptions made in all our algorithms. 

\textbf{Shift Mutation Strategy.} All algorithms described above use standard bit mutation as variation operator. To avoid sampling offspring that are identical to the parent (these offspring would not bring any new information to our optimization process, and are therefore useless), we use the ``shift'' operation suggested in~\cite{practice-aware}. If an offspring equals its parent, this strategy simply flips a randomly chosen bit. We write $y \gets \mutate(x,p)$ if $y$ is sampled by applying the shift mutation operator with mutation rate $p$ to $x$. 

\textbf{Termination Criterion and Runtime Measure.} 
We focus in this work on the \textit{runtime} (also known as \textit{optimization time}), which we measure in terms of generations that are needed until an optimal solution is evaluated for the first time. Since we only study algorithms with static offspring population size $\lambda$, the classical runtime in terms of function evaluations is easily obtained by multiplication with $\lambda$. As common in the academic benchmarking of EAs, our termination criterion is thus the state $f(x)=\max\{f(y) \mid y \in \{0,1\}^n\}$.  

\textbf{Strict vs. Non-Strict Update Rules.} We have presented in the previous section the algorithms as originally suggested in the literature. However, in our initial experiments we have made an interesting observation that the \oplab can substantially benefit from a slightly different parameter update rule, which replaces $p$ by $Ap$ only if the best offspring $y$ is \emph{strictly better} than the parent, i.e., if it satisfies $f(y)>f(x)$. We perform all experiments for the strict and the classical (non-strict) update rules, which -- together with the two lower bounds $p_{\min}=1/n^2$ and $p_{\min}=1/n$ -- yields four different settings for each benchmark problem. For reasons of space we can only comment on a few selected cases below. 
The detailed results are available in the appendix.
We mostly focus on the case of the strict update rule, if not stated otherwise.

\section{Hybridizing Q-Learning and the 1/5-th Success Rule}
\label{sec:qeas} 

The main contribution of our work is an algorithm that avoids the drawbacks of the above-mentioned \opl variants observed on \OneMax, and shows stable performance for all values of $\lambda$. We will achieve this by hybridizing the \oplab with Q-learning. 
%

\textbf{Q-learning} is a method that falls into the broader category of reinforcement learning (RL). Q-learning aims at learning, from the data that it observes, a policy that tells an \emph{agent} which \emph{action} to apply in a given situation. For this, it maintains a state-action matrix, in which it records its guess for what the expected \emph{reward} of each action in each of the states is. For a given \emph{state} $s$, the action $a$ maximizing this expected reward is chosen and executed. The environment returns a numerical reward and a representation of its state. The reward is used to update the state-action matrix, according to some rules that we shall discuss in the next paragraphs. The Q-learning process repeats until some termination criterion is met. The goal of the agent is to maximize the total reward. A smooth introduction to RL can be found in~\cite{sutton}. 

\textbf{The \qea.} We apply Q-learning to control the mutation rate of the \opl with fixed offspring population size $\lambda$. 
We first present in Alg.~\ref{alg:oplQ} the basic \qea. Its hybridization with the 1/5-th success rule will be explained further below. The \qea considers only two actions: whether to multiply the current mutation rate $p$ by the factor $A>1$ (action $a_\text{mult}$) or whether to multiply it by the factor $b<1$ (action $a_\text{divide}$). As mentioned in the introduction, the advantage of this action space is a smooth transition of the mutation rates between consecutive iterations, compared to a possibly abrupt change when operating directly on the parameter values. 

\begin{algorithm2e}[t]
    \textbf{Input:}{ population size $\lambda$, learning rate $\alpha$, learning factor $\gamma$}\;
     \textbf{Initialization:}\\
     \Indp
	{$x \gets \text{random string from } \{0,1\}^n$}\;
    {$p \gets 1/n$}\;
    \lFor{\highlight{all states $s_i \in [0\ldots\lambda]$ and all actions $a_i \in \{a_\text{mult}, a_\text{divide}\}$}}
        {\highlight{$Q(s_i, a_i) \gets 0$}}
    \highlight{$s, a \gets \text{undefined}$}\label{init-end}\;
    \Indm
\textbf{Optimization:} 
    \While{termination criterion not met}{              
        \lFor{$i= 1,\ldots,\lambda$}
            {$y^{(i)} \gets \mutate(x,p)$}
        {$x^* \gets \argmax_{y^{(i)}}{f(y^{(i)})}$}\; 
        {$x_\text{old} \gets x$}\; 
        \lIf{$f(x^*) \geq f(x)$}{$x \gets x^*$}\label{opl-end}
        \highlight{$r \gets \frac{f(x^*)}{f(x_\text{old})} - 1$} \tcp*{reward calculation}\label{reward} 
        \highlight{$s' \gets 0$}\label{state-start}\;
        \For{\highlight{$i =1,\ldots,\lambda$}}{ 
            \If{\highlight{$f(y^{(i)}) > f(x_\text{old})$}\label{line:ineq1}}
                {\highlight{$s' \gets s' + 1$}\label{state-end}  \tcp*{state calculation}}
                }
        \If{\highlight{$s \neq \text{undefined}$ \textbf{and} $a \neq \text{undefined}$} \label{undef}}
            {\highlight{$Q(s,a) \gets Q(s,a) + \alpha \left(r + \gamma \max_{a'}Q(s',a') - Q(s,a)\right)$;}\label{update}}
        \highlight{$s \gets s'$}\;            
        \eIf{\highlight{$Q(s', a_\text{mult}) = Q(s', a_\text{divide})$} \label{action-start}}
          {\highlight{$a \gets$ select $a_\text{mult}$ or $a_\text{divide}$ equiprobably; \label{replace}}}
            {\highlight{$a \gets \argmax_{a'}{Q(s', a')}$}\label{action-end}\;}
        {$p \gets ap$}\label{apply} \tcp*{update mutation rate} 
        {$p \gets \min(\max(p_\text{min}, p), p_\text{max})$}\label{bound} \tcp*{capping mutation rate}
    }
\caption{The \qea, \highlight{Q-learning highlighted in blue font}}
\label{alg:oplQ}  
\end{algorithm2e}

We use as reward the relative fitness gain, i.e., $(\max f(y^{(i)})-f(x))/f(x)$ (where we use the same notation as in the description of the \opl, i.e., $x$ denotes the parent individual and $y^{(1)}, \ldots, y^{(\lambda)}$ its $\lambda$ offspring). This reward is computed in line~\ref{reward}. Note here that several other reward definitions would have been possible. We tried different suggestions made in~\cite{Karafotias-rewards} and found this variant to be the most efficient.  
%
%
The new state $s'$ is computed as the number of offspring $y^{(i)}$ that are strictly better than the parent (lines~\ref{state-start}-\ref{state-end}). With the reward and the new state at hand, the efficiency estimation $Q(s,a)$ is updated in line~\ref{update}, through a standard Q-learning update rule. Note here that action $a$ is the one that was selected in the previous iteration (lines~\ref{action-start}-\ref{action-end}), and it resulted in moving from the previous state $s$ to the current state $s'$. 

After this update, the \qea selects the action to be used in the next iteration, through simple greedy selection if possible, and through an unbiased random choice otherwise; see lines~\ref{action-start}-\ref{action-end}. 
The mutation rate $p$ is then updated by this action (line~\ref{apply}) and capped to remain within the interval $[p_{\min},p_{\max}]$ if needed (line~\ref{bound}).

\emph{Hyperparameters.} The \qea has six hyperparameters, the constant factors of the actions $a_\text{mult}$ and $a_\text{divide}$, the upper and lower bounds for the mutation rate $p_{\min}$ and $p_{\max}$, and two hyperparameters originating from the Q-learning methodology itself (line~\ref{update}), the \textit{learning rate} $\alpha$ and the \textit{discount factor} $\gamma$. In our experiments, we use $a_\text{mult}=2$, $a_\text{divide}=1/2$, $p_\text{max} = 1/2$, $\alpha = 0.8$, and $\gamma = 0.2$. These values were chosen in a preliminary tuning step, details of which we have to leave for the full report due to space restrictions. For $p_{\min}$ we show results  for two different values, $1/n^2$ and $1/n$, just as we do for the other parameter control mechanisms.  

\textbf{The \hqea, the Hybrid Q-learning EA.} 
In the hybridized \qea, the \hqea, we reconsider the situation when the $Q(s,a)$ estimations are equal. This situation arises in two cases: when the state $s$ is visited for the first time or when the same estimation was learned for both actions $a_\text{mult}$ and $a_\text{divide}$. In these cases, the learning mechanism cannot decide which action is better, and an action is selected uniformly randomly. 
The \hqea, in contrast, borrows in this case the update rule from the \oplab algorithm, i.e., action $a_{\text{mult}}$ is selected if the best offspring is strictly better than the parent, otherwise $a_\text{divide}$ is chosen. Formally, we obtain the \hqea by replacing in Alg.~\ref{alg:oplQ} line~\ref{replace} by the following text: 
\begin{equation}\label{eq:ab-rule}
    \textbf{if }{f(x^*) > f(x_{\text{old}})} \textbf{ then } {a \gets a_\text{mult}} \textbf{ else } {a \gets a_\text{divide}.}
\end{equation}

\textbf{Strict vs. Non-Strict Update Rules.} As mentioned at the end of Sec.~\ref{sec:opl}, we experiment both with a strict and a non-strict update rule. Motivated by the better performance of the strict update rule, the description of the \qea and the \hqea use this rule. The non-strict update rules can be obtained from Alg.~\ref{alg:oplQ} by replacing the strict inequality in line~\ref{line:ineq1} by the non-strict one. Similarly, for the \hqea, we also replace ``\textbf{if}{$f(x^*) > f(x_{\text{old}})$}'' in (\ref{eq:ab-rule}) by ``\textbf{if}{$f(x^*) \ge f(x_{\text{old}})$}''. 
\section{Empirical comparison of parameter control algorithms}
\label{sec:results}

We now demonstrate that, despite the seemingly minor change, the \hqea outperforms both its origins, the \qea and the \oplab, on several benchmark problems. 
We recall that the starting point of our investigations were the results presented in~\cite{Rodionova-GECCO}, which showed that the performance of the \opl variants discussed in Sec.~\ref{sec:opl} on \onemax strongly depends on (1) the offspring population size $\lambda$, and on (2) the bound $p_{\min}$ at which we cap the mutation rate.  The \hqea, in contrast, is shown to yield stable performance for all tested values of $\lambda$ and for both tested values of $p_{\min}$.

\textbf{Experimental setup.} All results shown below are simulated from 100 independent runs of each algorithm. We report statistics for the optimization time, i.e., for the random variable counting the number of steps needed until an optimal solution is queried for the first time. Since the value of $\lambda$ is static, we report the optimization times as number of generations; classical running time in terms of function evaluations can be obtained from these values by multiplying with $\lambda$. For \onemax, we report average optimization times, for consistency with the results in~\cite{Rodionova-GECCO} and with theoretical results. However, for some of the other benchmark problems, the dispersion of the running times can be quite large, so that we report median values and interquartile ranges instead. Please also note that we use logarithmic scales in all runtime plots.

In the cases of large dispersion, we also performed the rank-sum Wilcoxon test to question statistical significance~\cite{stattest}. More precisely, we compared the \hqea to each of the other algorithms. As the input data for the test, the runtimes of all 100 runs of each of the two compared algorithms were used. The significance level was set to $p_0 = 0.01$.

The value of $\lambda$ is parameterized as $2^t$, with $t$ taking all integer values ranging from 0 to 12 for \onemax and from 0 to 9 for all other problems. The problem dimension, in contrast, is chosen in a case-by-case basis. We recall that it was shown in~\cite{Rodionova-GECCO} that the dimension did not have any influence on the ranking of the algorithms on \onemax. This behavior can be confirmed for the here-considered algorithm portfolio (results not shown due to space limitations).  

\subsection{Stable Performance on OneMax}
\label{sec:OM}

\begin{figure}[t]
  \centering
    \includegraphics{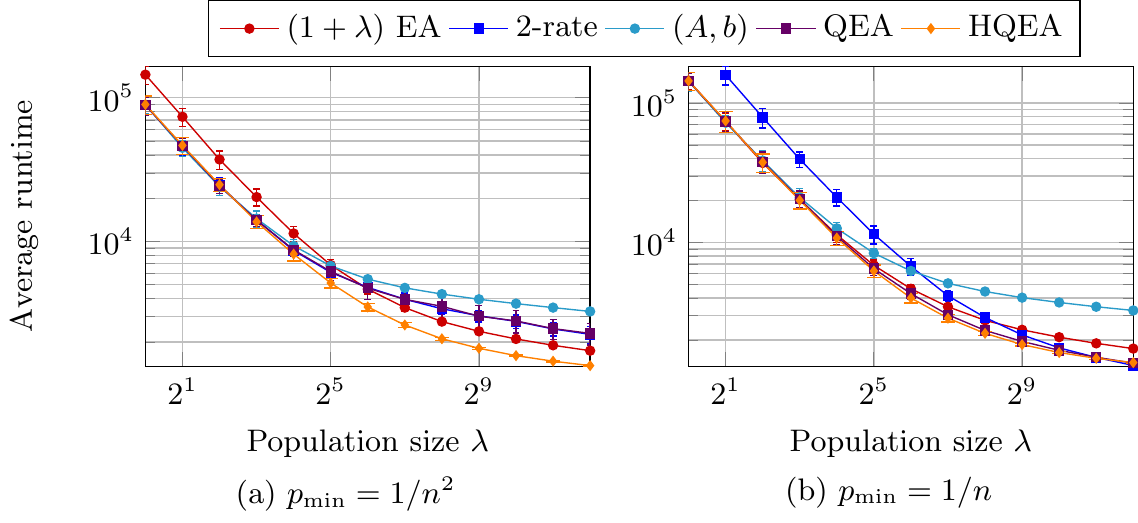}
   \caption{Average number of generations and its standard deviation needed to locate the optimum of the \OneMax problem}
  \label{fig:om-strict}
\end{figure}


Fig.~\ref{fig:om-strict} summarizes our empirical results for the $10^4$-dimensional \onemax problem, the problem of maximizing the function $\OM:\{0,1\}^n \to [0..n], x \mapsto \sum_{i=1}^n{x_i}$.   
For $p_{\min}=1/n^2$, our key findings can be summarized as follows. 
\textbf{(i)}~For small $\lambda$ up to $2^4,$ all the parameter control algorithms perform similarly and all of them seem to be significantly better than the \lea with static mutation rates.
\textbf{(ii)}~Starting from $\lambda > 2^5$ for the \AB and from $\lambda > 2^6$ for the \oplQ and the \rate, these algorithms are outperformed by the \lea.
\textbf{(iii)}~The \hybrid is the only parameter control algorithm that substantially improves the performance of the \lea for all considered values of $\lambda.$ The advantage varies from 21\% for $\lambda = 2^{12}$ to 38\% for $\lambda = 1$.

For the less generous $p_{\min}=1/n$ lower bound, we observe the following.
\textbf{(i)}~Overall, the performance is worsened compared to the $1/n^2$ lower bound. In particular, for small values of $\lambda,$ most of the algorithms are indistinguishable from the \lea, except for the \rate, which is even substantially worse.
\textbf{(ii)}~However, for $\lambda \geq 2^9,$ the \rate starts to outperform the \lea, in strong contrast to the situation for the $1/n^2$ lower bound.
\textbf{(iii)}~Our \hybrid is the only method which is never worse than the \lea and still outperforms it for $\lambda > 2^6$. 
With the growth of $\lambda$, the advantage grows as well:
while the \lea with $\lambda=2^{12}$ needs 1738 generations, on average, the \hybrid only requires 1379 generations, an advantage of more than 20\%.
\textbf{(iv)}~It is worth noting that the \oplQ in this case performs on par with the \hybrid.

Overall, we thus see that the \hybrid is the only considered parameter control algorithm, which stably performs on par or better than the \lea and all of the other algorithms for all values of $\lambda$ and for both values $p_{\min} \in \{1/n^2, 1/n\}$.

\subsection{Stable Performance on Other Benchmark Problems }

\textbf{LeadingOnes.} 
The \LeadingOnes problem asks to maximize functions of the type $\LO_{z,\sigma}:\{0,1\}^n \to \R, x \mapsto \max\{i \in [n] \mid \forall j \le i: x_{\sigma(i)}=z_{\sigma(i)}\}$, where $\sigma$ is simply a permutation of the indices $1,\ldots,n$ (the classic \LO
function uses the identity). We study the $n = 10^3$-dimensional variant of this problem. 

For $p_{\min}=1/n^2$ 
all the methods -- including the \lea -- show very similar performance, with the difference between the best and the worst of the five algorithms varying from 3\% to 6\% for each offspring population size $\lambda$, which is of the same order as the corresponding standard deviations. 
For the $1/n$ lower bound, the situation is similar, except that the \rate performs substantially worse than the \lea for all considered values of $\lambda,$ and the difference varies from 45\% to 93\%.

As a result, the \hybrid generally performs on par with the \lea for all considered values of $\lambda$ and both considered lower bounds on the mutation rate. Particularly, for $p_{\min}=1/n^2$ it is strictly better in 6 of the 10 cases, and in the other cases the disadvantages are 0.3\%, 0.3\%, 0.7\%, and 1.1\%. 

\textbf{Neutrality.}  
The \Neutrality function is a W-model transformation~\cite{Wmodel} that we apply to \OneMax. It is calculated the following way: a bit string $x$ is split into blocks of length $k$ each, and each block contributes 0 or 1 to the fitness value according to the majority of values within the block. In line with~\cite{Wmodel} and~\cite{DoerrYHWSB20} we considered $k = 3$. We study the $n = 10^3$-dimensional version of this problem. The results are summarized in Fig.~\ref{pic:neu}.

\begin{figure}[t]
\includegraphics[width=\textwidth]{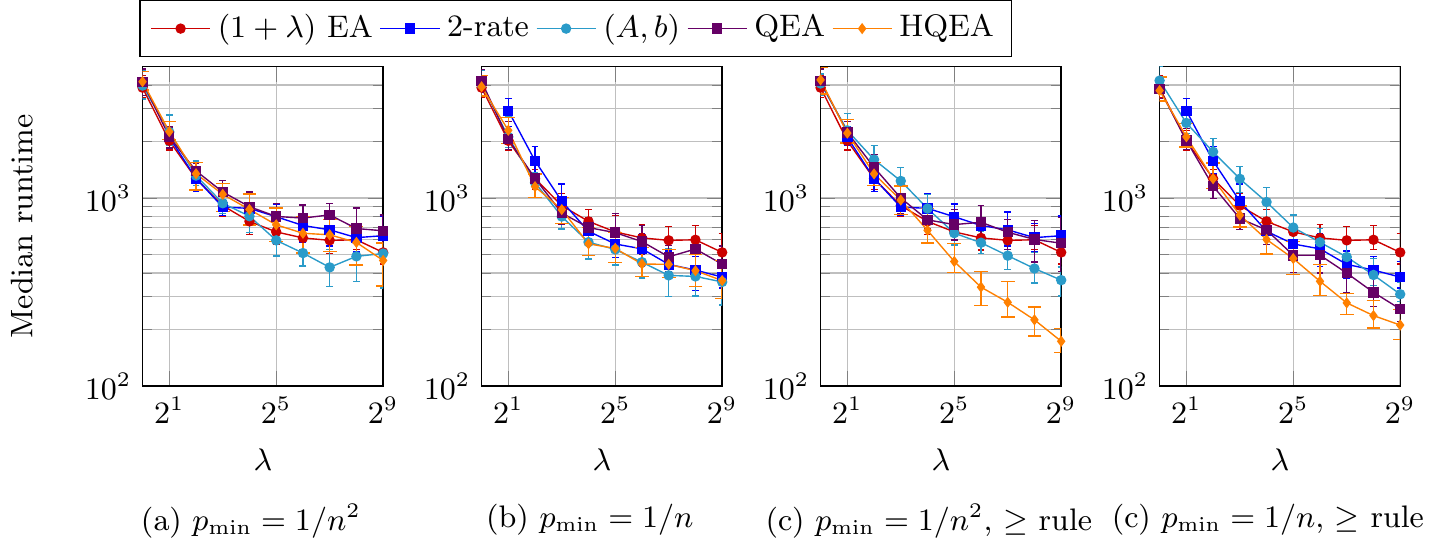}
\caption{Median number of generations and the corresponding interquartile ranges needed to locate the optimum of the \Neutrality problem}
\label{pic:neu}
\end{figure}

For $p_{\min}=1/n^2$ we obtain the following observations. 
Most of the parameter control methods perform poorly, i.e. worse than the \lea.
The exception is \AB, which performs better than the \lea for several values of $\lambda$ (in particular, $\lambda = 2^6, 2^7$). 

The lower bound $p_{\min}=1/n$ turns out to be preferable for all the algorithms: for large offspring population sizes $\lambda$, they all perform better than the standard \lea. Our \hybrid is usually one of the best algorithms, but however, for $\lambda = 2^7$ and $\lambda = 2^8$ it seems to be worse than the \AB. The Wilcoxon test results did not confirm the significance of this difference though (the p-values are greater than 0.04 in both cases).

For this problem we also observe that switching from the strict update rule to the non-strict version 
is beneficial for the \hybrid, the \oplQ, and the \AB, regardless of the value of $p_{\min}$. 
It is worth noting that with these values of hyper-parameters the \hybrid performs significantly better on high values of the population size ($\lambda \geq 2^5$) than all the other considered methods (the p-values are between $1.6 \cdot 10^{-9}$ and $3.9\cdot 10^{-18}$).

\textbf{Plateaus.} Plateau is an extension of the W-model suggested in~\cite{DoerrYHWSB20}. This transformation operates on the function values, by setting $\Plateau(f(x)) := \lfloor f(x)/k \rfloor + 1$, for a parameter $k$ that determines the size of the plateau. 
We superpose this transformation to \OneMax, and study performances for dimension $n = 1000$. 

\emph{Small plateaus, $k=2$.} 
For $k=2$, $p_{\min}=1/n^2$, and  $2 \leq \lambda \leq 2^6$, all considered parameter control algorithms improve the performance of the \lea. For large values of $\lambda$ (starting from $\lambda = 2^7$), however, the runtimes of the \lea and the parameter control algorithms are hardly distinguishable. The only exception for large $\lambda$ is the proposed \hybrid, which performs a bit better than the \lea. The Wilcoxon test suggests that the difference is significant with the p-values less than $3.9 \cdot 10^{-18}$.

The results obtained when using $p_{\min}=1/n$ are less successful, as most of the parameter control methods just perform on par with the \lea in this case. 
The \hybrid shows nevertheless a stable and comparatively good performance for all offspring population sizes $\lambda$. The \rate performs worse than the \lea in this case.

\emph{Plateaus with $k=3$.} We also considered a harder version of the problem with a larger size of the plateau, for which we use $k = 3$. As the total running time for this problem is much larger than for $k=2$, we had to restrict our experiments to a smaller problem size $n = 100.$

For $p_{\min}=1/n^2$ 
we cannot see any clear improvement of parameter control over the \lea any more. Moreover, for $\lambda \geq 2^7$, the \lea seems to be the best performing algorithm.

Interestingly, for the $1/n$ lower bound the situation is pretty similar to the $k = 2$ case. All the parameter control algorithms perform on par with the \lea (with only slight differences at $\lambda = 2^4, 2^6$), except for the \rate, which performs worse. 
It seems that as the problem gets harder, a larger lower bound is preferable, which seems to be natural, as with a bigger plateau, a higher mutation rate is needed to leave it. 
Let us also mention that the \hybrid performs stably well for all considered values of $\lambda$ in this preferable configuration. 

\textbf{Ruggedness.} We also considered the W-Model extension 
F9 from~\cite{DoerrYHWSB20}, which adds local optima to the fitness landscape by mapping the fitness values to $r_2(f(x)):=f(x)+1$ if $f(x) \equiv n \mod 2$ and $f(x)<n$, $r_2(f(x)):=\max\{f(x)-1,0\}$ for $f(x) \equiv n+1 \mod 2$ and $f(x)<n$, and $r_2(n):=n$. 
This transformation is superposed on \OneMax of size $n = 100$. 

For $p_{\min}=1/n^2$, all the considered parameter control algorithms significantly worsen the performance of the \lea. 
Even the \rate, which, untypically, performs the best among all these algorithms, is still significantly worse than the \lea.

The situation improves for $p_{\min}=1/n$ and the parameter control algorithms show similar performance as the \lea. 
The only exception is again \rate, whose performance did not change much compared to the case $p_{\min}=1/n^2$.


\section{Conclusions and Future Work}
\label{sec:summaryfindings}

To address the issue of unstable performance of several parameter control algorithms on different values of population size reported in~\cite{Rodionova-GECCO}, we proposed the Q-learning based parameter control algorithm, the \qea, and its hybridization with the \oplab, the \hqea. The algorithms were compared empirically on \OneMax and five more benchmark problems with different characteristics, such as neutrality, plateaus and presence of local optima. Our main findings may be summarized as follows.

On simple problems, i.e. \OneMax, \LeadingOnes, and \Plateau with $k =2$ the \hybrid is the only algorithm which always performs on par or better than the other tested algorithms for all the considered values of $\lambda$ and both mutation rate lower bounds.

On the harder problems, i.e., \Neutrality, \Plateau with $k = 3$, and \Ruggedness, the \hybrid performance depends on the lower bound (the same is true for the other algorithms). For $p_{\min}=1/n$, the \hybrid still performs on par with or better than the other algorithms for all values of $\lambda$ in almost all cases. 

The \oplQ is usually worse than the \hybrid. There are a number of examples where \AB is significantly worse as well. The hybridization of these two algorithms seems to be essential for the observed good performance of the \hybrid. 

As next steps, we plan on investigating \emph{more possible actions} for the Q-learning part. For example, one may use several different multiplicative update rules, to allow for a faster adaptation when the current rate is far from optimal. This might in particular be relevant in \emph{dynamic environments}, in which the fitness functions (and with it the optimal parameter values) change over time.  
We also plan on identifying ways to automatically select the configuration of the Q-learning algorithms, with respect to its hyper-parameters, but also with respect to whether to use the strict or the non-strict update rule. In this context, we are investigating exploratory landscape analysis~\cite{mersmann2011exploratory,Verel16HDR}. 

\vspace{1ex}
{\footnotesize{
\textbf{Acknowledgments.} The reported study was funded by RFBR and CNRS, project number 20-51-15009, by the Paris Ile-de-France Region, and by a public grant as part of the
Investissement d'avenir project, reference ANR-11-LABX-0056-LMH,
LabEx LMH. 
}}

}

\newcommand{\showDOI}[1]{\skip}
\newcommand{\showaddress}[1]{\skip}

%
%
\appendix
\section*{Appendix}

\centering

    \includegraphics[width=\textwidth]{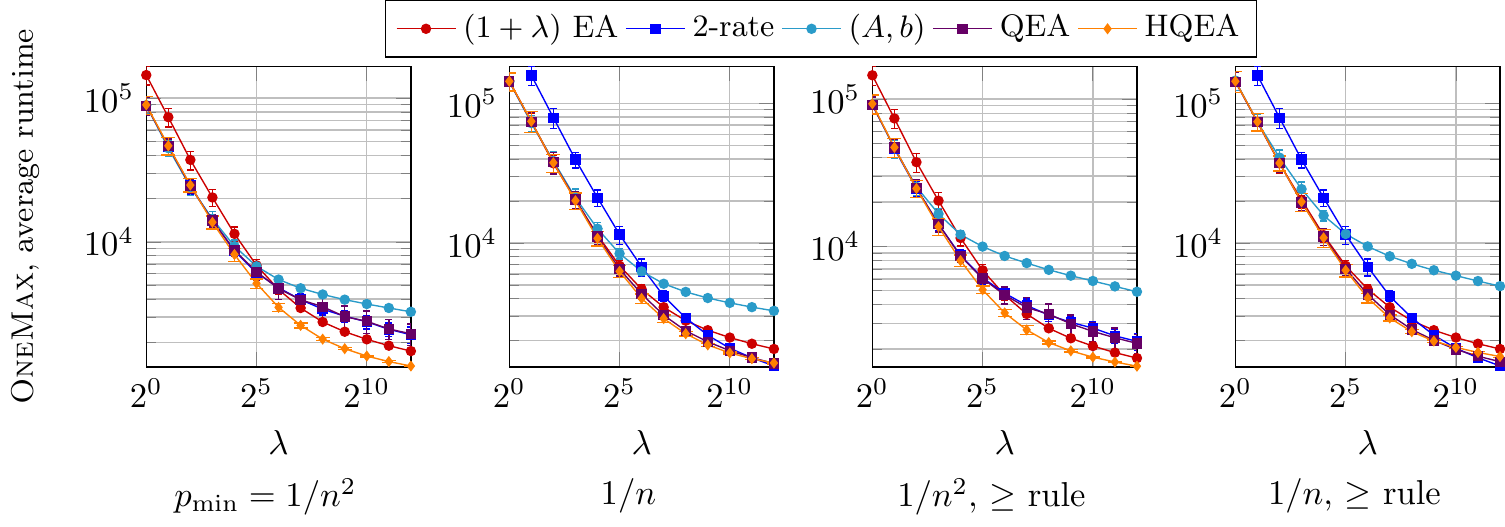}

    \includegraphics[width=\textwidth]{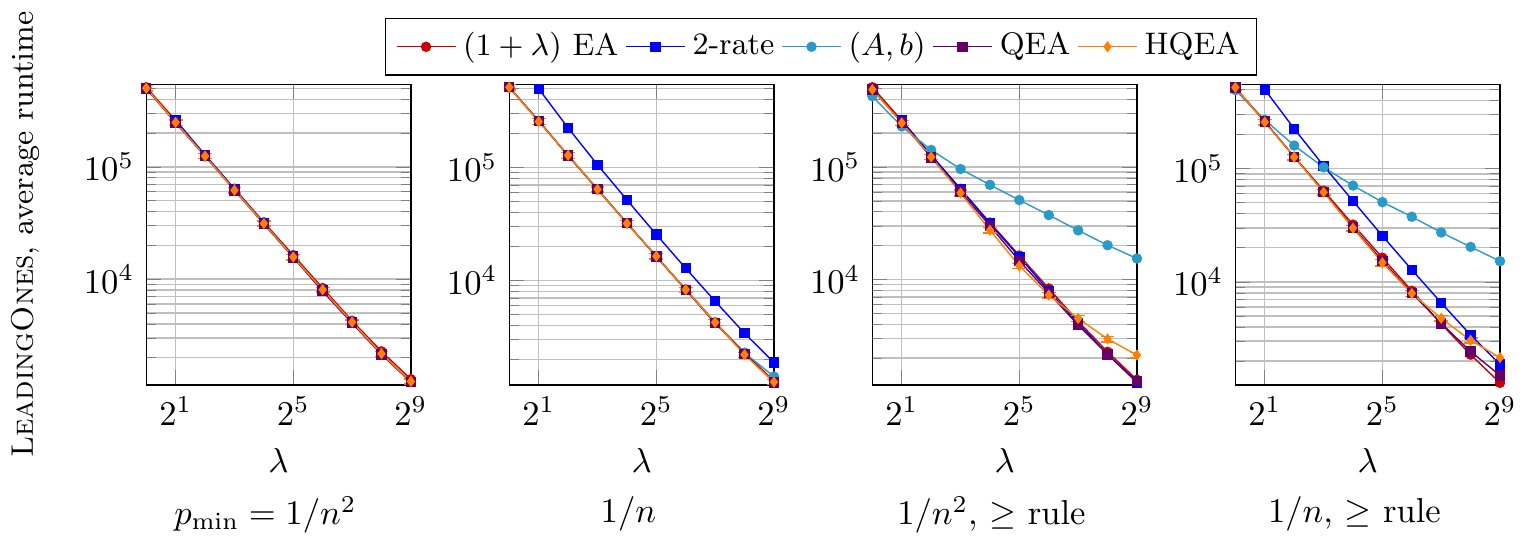}

    \includegraphics[width=\textwidth]{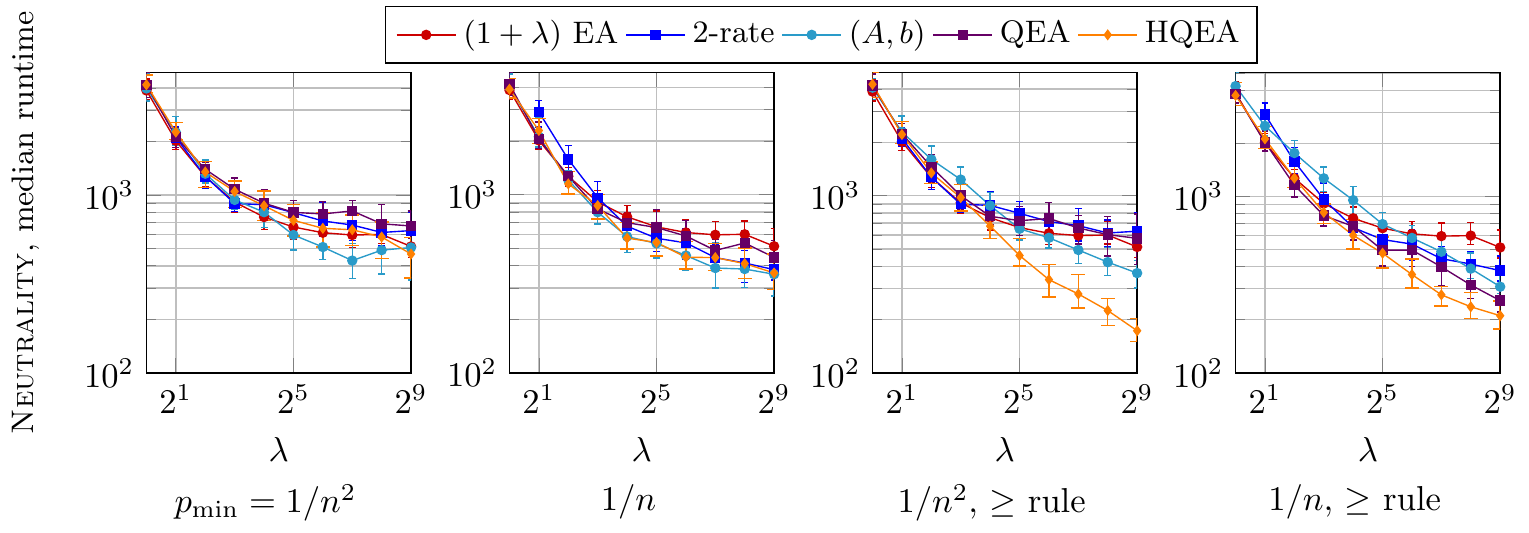}

    \includegraphics[width=\textwidth]{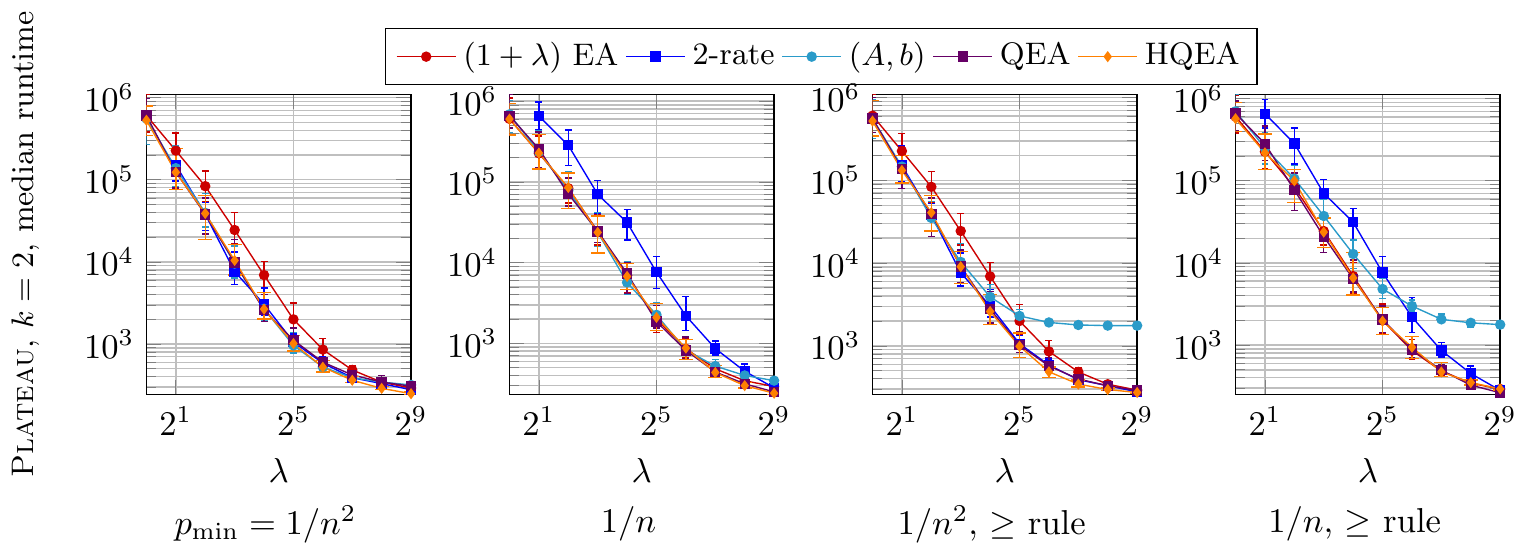}

    \includegraphics[width=\textwidth]{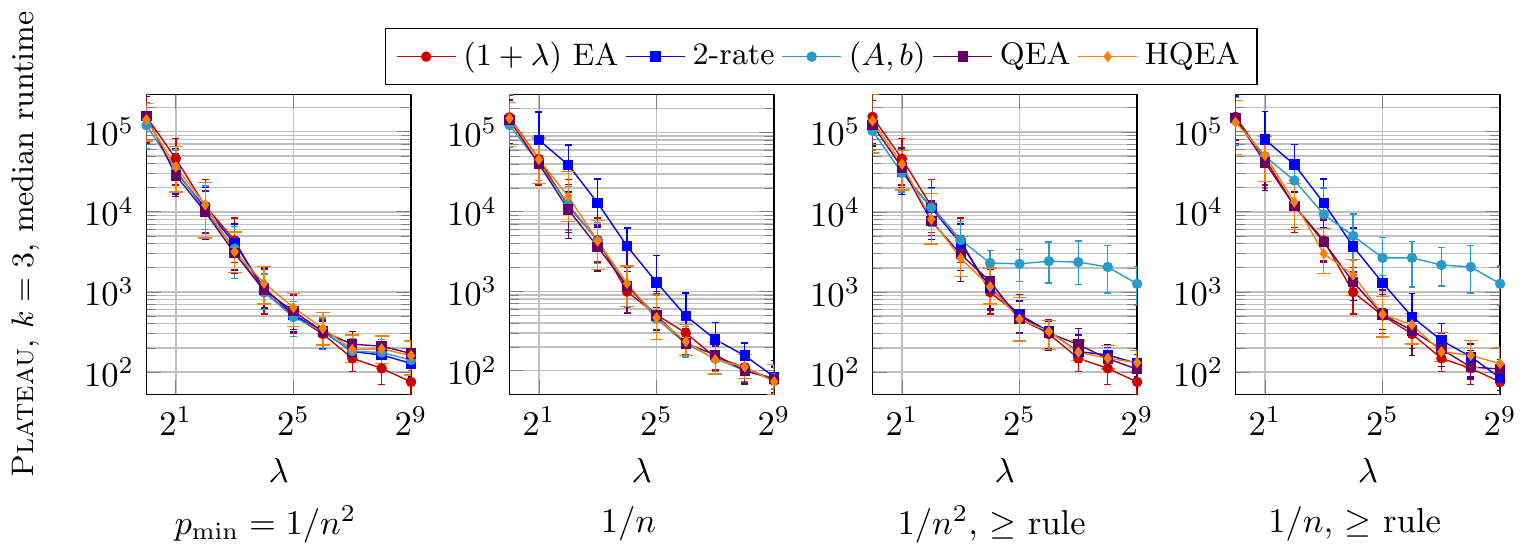}

    \includegraphics[width=\textwidth]{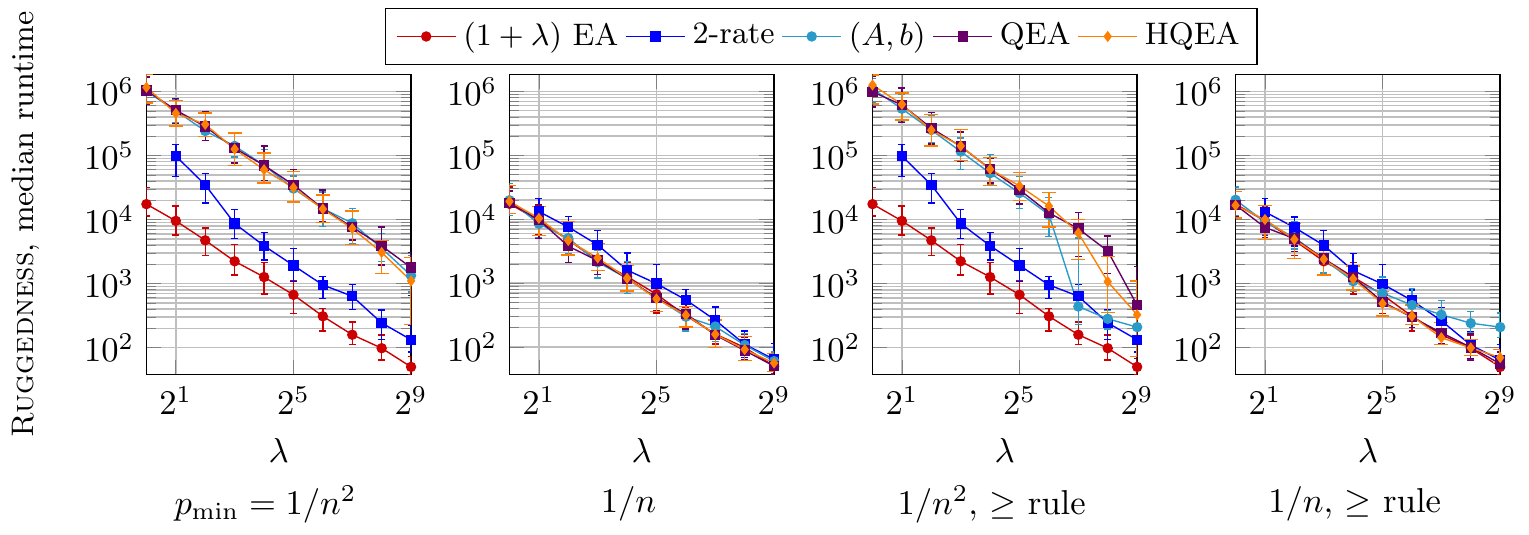}

\end{document}